\documentclass[runningheads]{llncs}
\usepackage{graphicx}
\usepackage{amsmath}
\usepackage{amssymb}
\usepackage{booktabs}
\usepackage{algorithm}
\usepackage{algpseudocode}
\usepackage{array}
\usepackage{xcolor}
\usepackage{ulem}
\usepackage{hyperref}
\usepackage{subcaption}
\usepackage{makecell}
\usepackage{appendix}

\usepackage{comment}
\excludecomment{appendix}

\newcolumntype{C}[1]{>{\centering\arraybackslash}p{#1}}

\begin{document}
\title{FUNAvg: Federated Uncertainty Weighted Averaging for Datasets with Diverse Labels}
\titlerunning{FUNAvg}
%
\author{Malte Tölle\inst{1,2,3,4}\orcidID{0000-0002-0804-5794} \and
Fernando Navarro\inst{4}\orcidID{0000-0001-8906-9079} \and
Sebastian Eble\inst{1}\orcidID{0009-0000-3736-4316} \and
Ivo Wolf\inst{1,5}\orcidID{0000-0002-6519-6484} \and
Bjoern Menze\inst{4}\thanks{These authors contributed equally.}\orcidID{0000-0003-4136-5690} \and
Sandy Engelhardt\inst{1,2,3}\textsuperscript{*}\orcidID{0000-0001-8816-7654}}
\authorrunning{M. Toelle et al.}
%
\institute{Department of Internal Medicine III, Heidelberg University Hospital, Germany \and
Informatics for Life Institute, Ruprecht-Karls University Heidelberg, Germany \and
DZHK (German Centre for Cardiovascular Research), partner site Heidelberg/Mannheim \and
Department of Quantitative Biomedicine, University of Zurich, Switzerland \and
Department of Computer Science, Mannheim University of Applied Sciences, Germany\\
\email{malte.toelle@med.uni-heidelberg.de}}
\maketitle              
%
\begin{abstract}
    Federated learning is one popular paradigm to train a joint model in a distributed, privacy-preserving environment. 
    But partial annotations pose an obstacle meaning that categories of labels are heterogeneous over clients.
    We propose to learn a joint backbone in a federated manner, while each site receives its own multi-label segmentation head.
    By using Bayesian techniques we observe that the different segmentation heads although only trained on the individual client's labels also learn information about the other labels not present at the respective site. 
    This information is encoded in their predictive uncertainty.
    To obtain a final prediction we leverage this uncertainty and perform a weighted averaging of the ensemble of distributed segmentation heads, which allows us to segment "locally unknown" structures.
    With our method, which we refer to as FUNAvg, we are even on-par with the models trained and tested on the same dataset on average. 
    The code is publicly available\footnote{\url{https://github.com/Cardio-AI/FUNAvg}}.
\end{abstract}

\section{Introduction}

The ability of deep neural networks to accurately segment anatomical structures promises precise quantitative analysis and clinical diagnosis, and has the potential to significantly improve medical decision making~\cite{ulrich2023multitalent}. 
However, due to the required effort for accurately labelling medical images every institution only creates annotations for their particular research endeavour~\cite{Radsch2023labeling}.
Despite being from the same modality and capturing a joint field of view, the cumulative information is hardly exploited.
This leads to partially annotated, distributed datasets in a sense that some structures might be segmented in one dataset (e.g. liver, kidney) and other structures in the other dataset (e.g., spleen, spine) that can not be gathered on a central server because of privacy constraints. 

\begin{figure}[t]
   \centering
   \includegraphics[width=\linewidth]{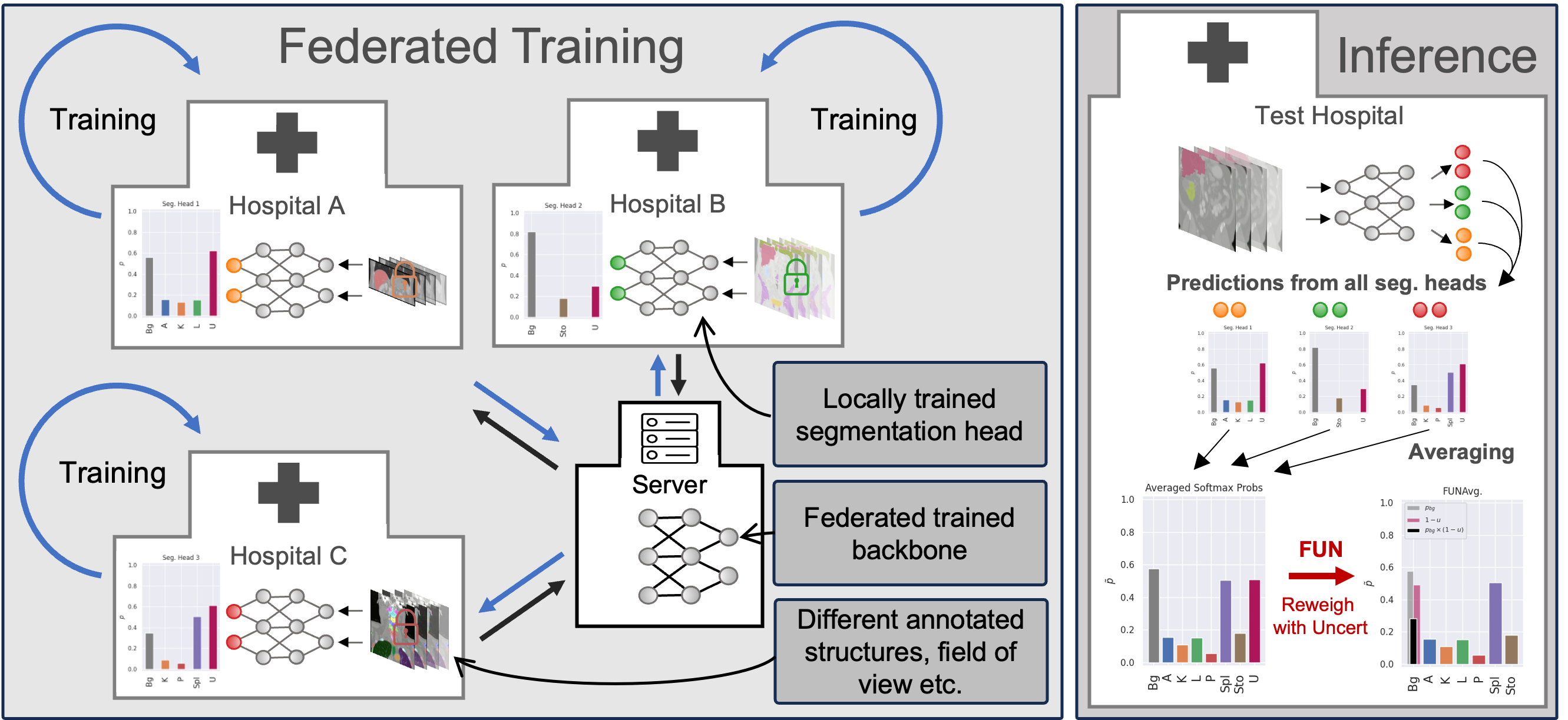}
   \caption{
   Our proposed training and inference scheme.
   During training each site optimizes its own segmentation head according to the number of present labels at the respective site.
   On the central server only the backbone is averaged in a federated fashion.
   During inference all segmentation heads are gathered and an average of the softmax probabilities is computed weighted by the number of sites the individual label was present.
   By utilizing the predictive uncertainty of the classifiers with FUNAvg the predictions can be improved, which is especially benefitial for underrepresented labels across the federated sites.
   }
   \label{fig:schematic-overview}
\end{figure}

The standard workflow includes training a model on dataset A that has the structure of interest annotated and applying it to dataset B, which due to domain gaps leads to subpar performance.
Recently, self-supervised methods have gained attention which use an unsupervised pre-training on a large corpus of data similar to dataset B, where the model is usually trained to minimize a reconstruction loss~\cite{hatamizadeh2021swinunetr}. 
Federated Learning (FL) is one renowned method that reverts the common paradigm of central data storage to circumvent privacy constraints~\cite{McMahan2017FedAVG}.
In FL the model is sent to each data holding institution and subsequently averaged on a central server during model training.
Our method combines both approaches by jointly learning all annotated structures across a multitude of datasets while showing that not-annotated information in the data is learned as well and therefore can be explored for improving predictions during inference similar to self-supervised methods.
The approach most related to ours, MultiTalent~\cite{ulrich2023multitalent}, also employs training multiple segmentation heads across different datasets but discards the possible information learned from not-annotated structures, partly because they do not employ a Bayesian approach~\cite{ulrich2023multitalent}.

We train one model in a federated setting across several partially labelled clients, which have different number of pre-segmented images and structures as well as annotation protocols.
Our method trains a backbone in a federated manner across all clients, while each client obtains a separate segmentation head.
The final prediction is then obtained by averaging in an ensemble-like manner.
Suprisingly, also those structures are represented in the uncertainty maps for models trained at sites where the corresponding labels were not available. 
We show that this uncertainty can be leveraged to segment previously unknown structures at this location to improve performance especially for structures that are underrepresented in the training sets i.e. are only present at one or few clients.

\begin{figure}[t]
    \centering
    \begin{subfigure}{\textwidth}
        \centering
        \includegraphics[width=\linewidth]{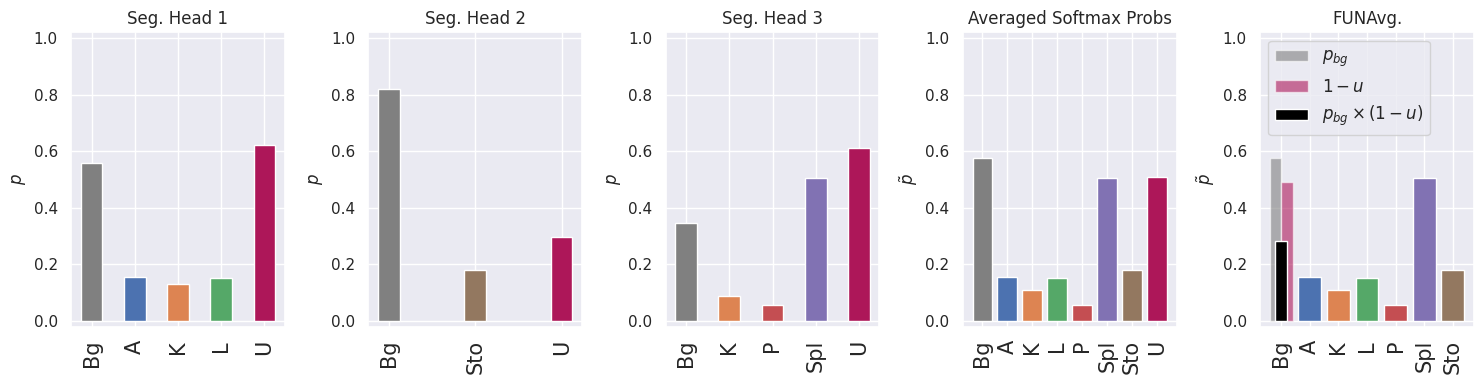}
        \caption{}
    \end{subfigure}
    \begin{subfigure}{\textwidth}
        \includegraphics[width=\linewidth]{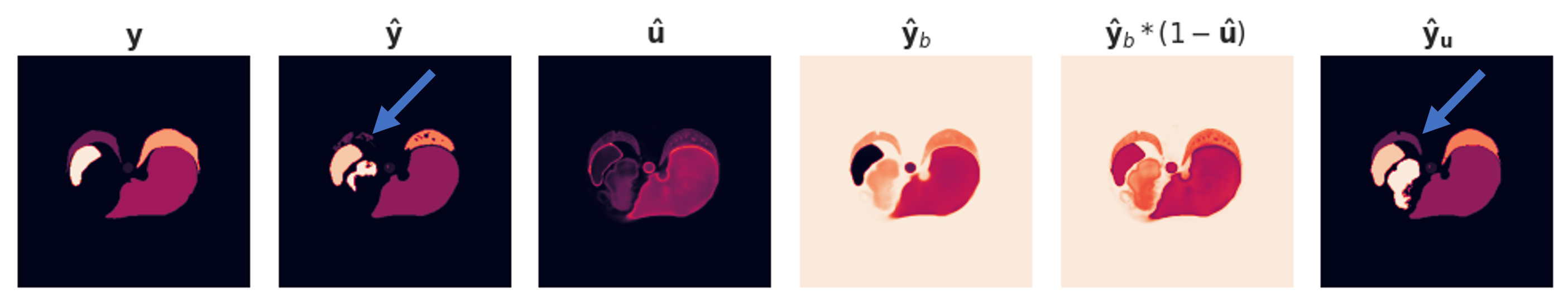}
        \caption{}
    \end{subfigure}
    \caption{
    Proposed federated uncertainty weighted averaging for (a) one pixel and (b) an entire image. 
    (a) When averaging the softmax probabilities of sites 1-3 the final prediction would be "Background" (Bg).
    By reweighting the probability for background by the uncertainty (U) we obtain the right label of "Spleen" in this example.
    (b) After averaging the logits the final prediction $\hat{y}$ is fragmentary especially in the area of the lung in above example, while the lung is perfectly visible in the uncertainty estimation $\hat{u}$. 
    We therefore multiply the background channel with the inverse of the uncertainty $\hat{y}_b \times (1-\hat{u})$ to obtain $\hat{y}_u$. 
    }
    \label{fig:uncert-avg}
\end{figure}

\section{Related Work}

\paragraph{Learning from Partially Annotated Datasets.}

Lately, there has been an increasing amount of studies focusing on multi-organ segmentation using data that is partially labeled. 
One line of research leverages the inherent homogeneous anatomy of the human body in terms of shape, size, and locations of anatomical structures~\cite{oktay2018acn,zhou2019pann}.
In another approach, the network is equipped with different encoders for each dataset in the learning procedure~\cite{xu2023fedmenu}.
But this method falls short to account for different combinations of the various organs across all datasets present in the model training.
Similar to equipping the model with different encoders multiple heads that share a custom backbone can be used~\cite{fuchs22multiheaduncert}.
In the approach most similar to ours called MultiTalent the network is equipped with different segmentation heads for each datasets in the learning procedure~\cite{ulrich2023multitalent}.
It is argued that this is largely needed due to different annotation protocols across the different datasets.
However, this fails at leveraging unannotated organs across different datasets that enhances predictive quality due to larger amount of data samples.
Further, we believe that varying annotation protocols represent some form of data dependent i.e. aleatoric uncertainty that can be captured with Bayesian techniques~\cite{Gal2016Diss,kendall2017uncert,kohl2018probunet}.
MultiTalent does not exploit ways of obtaining a final prediction by e.g. averaging of the trained segmentation heads~\cite{ulrich2023multitalent}.

\begin{figure}[t]
    \centering
    \includegraphics[width=\linewidth]{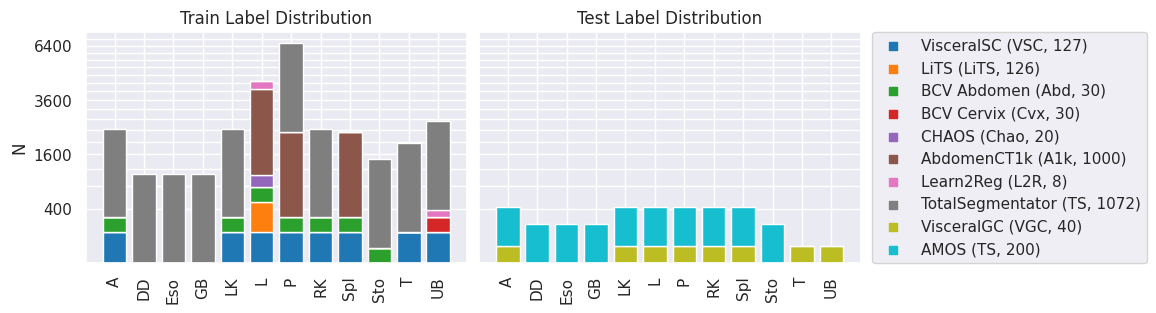}
    \caption{
    Datasets used for training and testing and their respective label distribution. 
    They differ in general quantity of training samples as well as number of annotated labels present. 
    We used the following open data: Liver Tumor Segmentation (LiTS)~\cite{bilic2023lits}, Beyond the Cranial Vault (BCV, Cervix and Abdomen)~\cite{landman2015bcv}, Combined Healthy Abdominal Organ (CHAOS)~\cite{kavur2019chaos}, Learn2Reg~\cite{zhoubing2016l2r}, AbdomenCT-1k~\cite{ma2022AbCT1k}, Abdominal Multi-Organ Benchmark (AMOS)~\cite{yuanfeng2022amos}, and TotalSegmentator~\cite{wasserthal2022totalsegmentator}, and two in-house datasets termed VisceralGC and SC.}
    \label{fig:dataset-overview}
\end{figure}

\paragraph{Bayesian Federated Learning.} 

In contrast to the point estimate used in frequentist deep learning, the Bayesian approach models the parameters with a distribution. 
Mathematically a prior distribution $p(\boldsymbol{\theta}|\alpha)$ is placed over the weights $\boldsymbol{\theta}$ of a neural network, governed by a hyperparameter $\alpha$.
Our interest lies in the posterior after observing some data $\mathcal{D}$ $p(\boldsymbol{\theta} | \mathcal{D},\alpha) = p (\mathcal{D}|\boldsymbol{\theta},\alpha) p(\boldsymbol{\theta}|\alpha) / p(\mathcal{D})$.
Unfortunately, this distribution is not tractable, but can be approximated with Bayesian inference techniques such as variational inference (VI).

In federated learning, numerous clients collaboratively train a unified global model, while adhering to data privacy as no data is shared with a central server managing the training process called federated averaging (FedAvg)~\cite{McMahan2017FedAVG}.
For a given set of clients, let $\mathcal{D}_i = \{(\boldsymbol{x}_i, \boldsymbol{y}_i)\}$, represent the data of client i and $\boldsymbol{\overline{\theta}}$ the weights of the global model. 
At the beginning of each training round the model weights $\boldsymbol{\theta_i}$ of every client are initialized with the latest global model weights. 
Subsequently, each client performs stochastic gradient descent.
After receiving the trained models from each client, FedAvg updates the global model by averaging the weights. 
This averaging is performed in proportion to the amount of data each client contributes, ensuring that clients with more data have a correspondingly greater influence on the final model $\boldsymbol{\overline{\theta}} \leftarrow \sum_i \frac{|n_i|}{n}\boldsymbol{\theta_i}$.
The final model results from the  iterative application of the two previously described equations. 

The standard approach of federated averaging has been refined for frequentist model training in mutiple works~(e.g., \cite{Karimireddy2020Scaffold,Li2020FedProx}).
A Bayesian model capable of predicting uncertainty can be obtained by training in the frequentist manner on each client and subsequently by treating each client model as a sample from the posterior~\cite{boughorbel2019fluncert,linsner2021uncertainty,thorgeirsson2021bfl}.
As FL includes a set of candidate client models it naturally lends itself towards model ensembling which is used in FedBE~\cite{Chen2021FedBE}.
While especially model ensembles have received significant attention in the community the usage of full Bayesian neural networks with variational inference in the federated setting has been rarely explored.

\section{Methods}

\paragraph{Uncertainty.}

To reduce computational load we opt for the practical Bayesian technique of Monte Carlo (MC) dropout~\cite{Gal2016Diss}.
Using dropout during training and testing has been shown to approximate the true posterior, while not introducing more computational load to capture uncertainty.
Uncertainty is divided into aleatoric (data dependent) and epistemic (model dependent) uncertainty.
We follow \cite{kwon2018uncertainty} and define uncertainty as

\begin{equation}
    \mathcal{U} = \underbrace{\frac{1}{T} \left( \sum_{t=1}^{T}  \mathrm{diag}(\hat{\mathbf{p}}_t) - \hat{\mathbf{p}}^{\otimes 2} \right)}_{\mathrm{aleatoric}} + \underbrace{\frac{1}{T} \sum_{t=1}^{T} \left( \hat{\mathbf{p}}_t - \overline{\mathbf{p}} \right)^{\otimes 2}}_{\mathrm{epistemic}} \;,
\end{equation}

\noindent where T is the number of MC sampling steps, $\otimes$ denotes the outer product, $f$ the network with parameters $\hat{\theta}_t$ in step $t$ for input $x^*$, $\overline{\mathbf{p}} = 1 / T \sum_{t=1}^{T} \hat{\mathbf{p}}_t$ and $\hat{\mathbf{p}}_t = \mathbf{p}(\hat{\boldsymbol{\theta}}_t) = \mathrm{Softmax}\{f^{\hat{\boldsymbol{\theta}}_t}(\mathbf{x}^{*})\}$.
The first part captures the inherent variability in the data that cannot be reduced even if more data were to be collected, while the second term captures the variance in the model's output that can potentially be reduced with more data.

Neural networks tend to produce overconfident predictions, meaning the predictive uncertainty is smaller than the predictive error, this is also known as miscalibration \cite{laves2022potobim} and can be expressed with

\begin{equation} \label{eq:miscalibration}
    \mathbb{E}_{\hat{p}} \left[ |\mathbb{P} \left( \hat{y} = y | \hat{p} = q \right) -q | \right] \;, \quad \forall q \in [0, 1] \;,
\end{equation}

\noindent which quantifys the expectation of the difference between predicted softmax likelihood $\hat{p}$ and accuracy and can be approximated by the Expected Calibration Error (ECE)~\cite{guo17a}.
The predicted probabilities from a neural network are partitioned into $M$ bins and a weighted average of the difference between accuracy and confidence (i.e. softmax likelihood) is taken:

\begin{equation}
    \mathrm{ECE} = \sum_{m=1}^{M} \frac{|B_m|}{n} | \mathrm{acc}(B_m) - \mathrm{conf}(B_m) | \;,
\end{equation}

\noindent with total number of inputs $n$ and set of indices $B_m$ whose confidences fall into that particular bin.
Intuitively for all pixels that are predicted with a softmax likelihood of 0.2 the expected accuracy should equal 20 percent~\cite{laves2022potobim}.

\paragraph{Uncertainty Weighted Averaging.}

For training a model across partially labelled clients each client's model gains a separate segmentation head i.e. the last layer. 
The backbone of the model until the last layer is trained in a federated fashion. 
This enables us to learn each client's labels while simultaneously perform feature extraction across all tasks similar to~\cite{ulrich2023multitalent}.
To obtain the final prediction we average the predictions from all classifiers per channel and divide each channel by the number of clients possessing annotations of that particular label $\overline{\mathbf{p}} = \frac{1}{\mathbf{k}} \sum_{k=1}^{K} \hat{\mathbf{p}}_k$ with $\mathbf{k} =[k_1, k_2,...] \in \mathbb{R}^{K}$  the number of clients that particular label was present and $\hat{\mathbf{p}}_k \in \mathbb{R}^{k_i \times H \times W}$ the predicted per-channel softmax probability of each segmentation head.

Surprisingly, we noticed that although all classifiers are not aware of at least two labels present only at other clients, these unannotated structures in their local dataset were visible in the classifier's predictive uncertainty (see Fig.~\ref{fig:qualitative-results}).
The segmentation heads seem to have learned that there might be a structure but did not have the ability to assign a specific label to it due to predefined channels.
We interpret the uncertainty as the probability of "something" being present, while the background represents the probability for "nothing".
Consequently, we reduce the likelihood for a pixel being background by the probability of the presence of a structure encoded in the uncertainty. 
In areas where $\mathbf{u}$ is large, the probability for a structure being present is high.
We thus multiply the background channel (the first channel 0) with the inverse of the uncertainty $\overline{\mathbf{p}}_0 = (1-\mathbf{u}) \overline{\mathbf{p}}_0$. 
Fig.~\ref{fig:uncert-avg} shows a graphical explanation of the proposed averaging procedure for one specific pixel and for the whole image.
Our approach represents a self-supervised method for reweighting the predictive probability by the self-supervised learned un-annotated structures for each segmentation head, while still obtaining an unambiguous prediction due to the preservation of probabilities.
To ensure the probabilities for each pixel sum to one we must adjust the MultiTalent pipeline to be trained with cross entropy followed by a softmax. 
This enables us for each pixel to have the softmax probability of "something" compared to the probability to "nothing" that sums to one.
We control for calibration of our probabilities in terms of ECE.

\paragraph{Data and Model Architecture.}
In total we used 8 datasets for training comprising a total of 2413 3D images across 12 unique classes and 2 datasets for testing with 240 samples.
The training datasets are further split 80/20\,\% for training and testing.
Each dataset represents one client in the federation.
We chose our test set such that no client is in possession of all labels and that each label is at least present once in both test- and training sets.
A notable characteristic of these datasets is their diversity in field of view, contrast and noise (see Fig.~6 in the appendix). 
This variation not only mirrors the real-world scenarios encountered in medical imaging but also challenges and validates the robustness of our model in handling diverse imaging conditions and patient-specific variations.
In Fig.~\ref{fig:dataset-overview} we show all labels present in the datasets and their respective distribution across the federation.
Preprocessing and training was done using the nnUNet framework~\cite{isensee2021nnunet} as in~\cite{ulrich2023multitalent} with the above mentioned adaptations.

\begin{table*}[t]
    \centering
    \caption{
    Comparison of DICE scores of trained models on unseen 20\% test splits and completely unseen sites. 
    For comparison we train a model on one dataset only and evaluated on the same and all others respectively (row 1 and 2).
    Further we compare our federated trained version to the centralized case (Fed Avg and Cen Avg).
    For both we also use uncertainty weighted averaging (FUNAvg and CUNAvg).
    \textbf{M} denotes the row-wise mean. 
    We applied a Wilcoxon signed-rank test for each dataset between vanilla and uncertainty weighted averaging.
    }
    \begin{tabular}{
      p{0.13\textwidth} |
      C{0.07\textwidth}
      C{0.07\textwidth} 
      C{0.07\textwidth} 
      C{0.07\textwidth} 
      C{0.07\textwidth} 
      C{0.07\textwidth} 
      C{0.07\textwidth} 
      C{0.07\textwidth} |
      C{0.07\textwidth} 
      C{0.07\textwidth} |
      C{0.07\textwidth}
    }
    \toprule
    \textbf{Method}  & \textbf{VSC} & \textbf{LiTS} & \textbf{Abd} & \textbf{Cvx} & \textbf{Chao} & \textbf{A1k} & \textbf{L2R} & \textbf{TS} & \textbf{VGC} & \textbf{Amo} & \textbf{M} \\
    \midrule
    Same & \textbf{76.92} & \textbf{95.21} & 85.83 & 85.87 & \textbf{97.68} & \textbf{90.02} & 74.36 & 82.29 & 78.84 & \textbf{80.86} & \textbf{84.79} \\
    Others & 63.76 & 41.63 & 64.10 & 42.54 & 37.11 & 37.06 & 27.81 & 67.14 & 57.09 & 58.88 & 49.71 \\
    \midrule

    Cen Avg & 77.52 & 91.24 & 84.56 & \textbf{85.05} & 95.41 & 82.98 & \textbf{87.95} & 69.65 & \textbf{87.44} & 55.57 & 81.74 \\
    CUNAvg & 74.51 & 87.95 & 82.39 & 80.4 & 92.86 & 80.16 & 85.91 & 69.07 & 83.18 & 57.60 & 79.40 \\[-0.5em]
    & \tiny $<0.01$ & \tiny $<0.01$ & \tiny 0.02 & \tiny 0.07 & \tiny 0.03 & \tiny 0.01 & \tiny 0.41 & \tiny 0.35 & \tiny $<0.01$ & \tiny 0.63& \tiny $<0.01$ \\
    Fed Avg & \textbf{77.67} & 92.51 & 84.22 & 78.25 & 91.92 & 87.73 & 85.11 & 77.94 & 85.12 & 65.94 & 82.64 \\
    FUNAvg & 77.26 & \textbf{93.06} & \textbf{86.87} & 78.88 & \textbf{94.63} & \textbf{86.63} & 86.35 & \textbf{85.19} & 85.31 & \textbf{73.49} & \textbf{84.77} \\[-0.5em]
    & \tiny 0.04 & \tiny 0.45 & \tiny 0.01 & \tiny 0.89 & \tiny 0.19 & \tiny 0.54 & \tiny 0.78 & \tiny 0.03 & \tiny 0.8 & \tiny $<0.01$& \tiny 0.07 \\
    \bottomrule
    \end{tabular}
    \label{tab:results}
\end{table*}

\section{Results}

We performed an 80-20\% train-test split at each client. 
For a baseline, we trained models on single clients (no FL) and evaluated them in the intra-client scenario on the same client (row 1 in Tab. \ref{tab:results}) and inter-client scenario on the test split of all other clients where the target label was available (row 2 in Tab. \ref{tab:results}). 
We then performed training similar to MultiTalent~\cite{ulrich2023multitalent}, where each dataset obtained a separate segmentation head. 
We performed vanilla averaging and uncertainty weighted averaging for the centralized (Cen Avg and CUNAvg, rows 3 and 4) and federated setting (Fed Avg and FUNAvg, rows 5 and 6). 

\begin{figure}[t]
    \centering
    \begin{subfigure}{.36\textwidth}
        \centering
        \includegraphics[width=\linewidth]{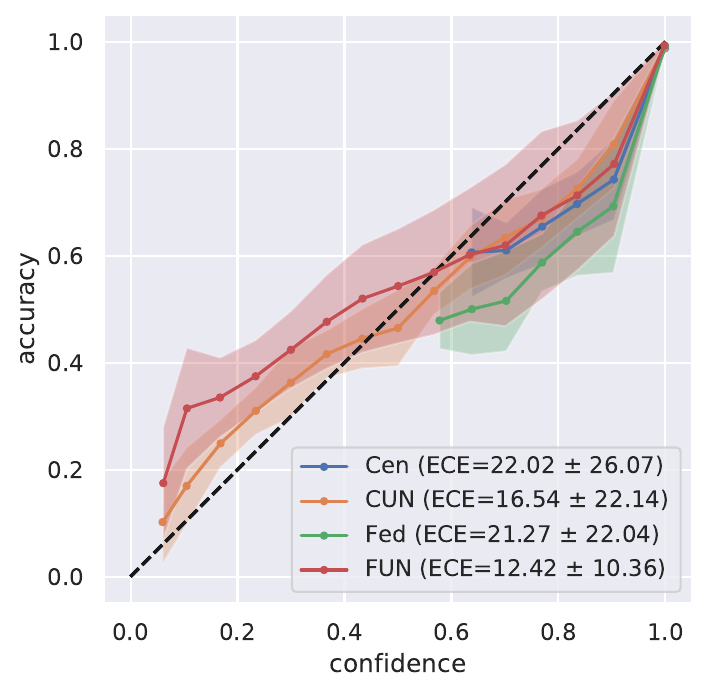}
        \caption{}
    \end{subfigure}
    \begin{subfigure}{.62\textwidth}
        \centering
        \includegraphics[width=\linewidth]{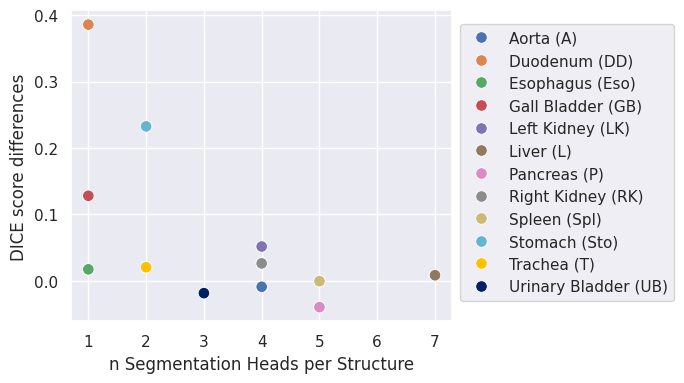}
        \caption{}
    \end{subfigure}
    \caption{Calibration of the different methods in terms of Expected Calibration Error (a) and performance gain of FUNAvg in comparison to vanilla averaging the different logits from the federated training segmentation heads (b).}
    \label{fig:ece-and-improvement}

    \centering
    \includegraphics[width=\textwidth]{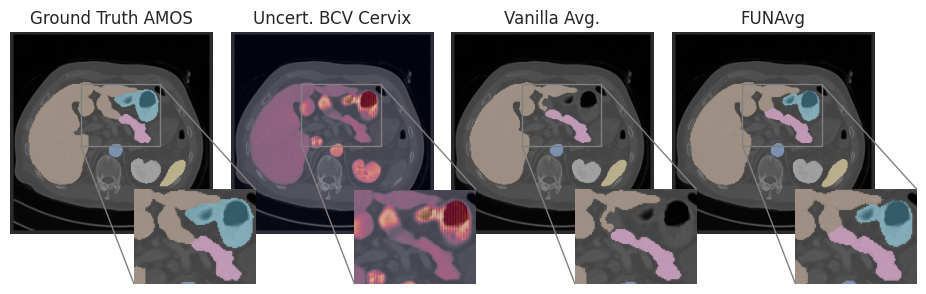}
    \caption{Qualitative results on the AMOS test set. The classifier trained on the BCV Cervix dataset does not predict any of the labels present in the ground truth. Still, it learns their presence in the uncertainty.}
    \label{fig:qualitative-results}
\end{figure}

Despite the challenges of different quantities of annotated labels, CT scans, and fields of view, federated training was successful for all structures present at any client. 
Our method (DICE=84.77) is on par with the models trained and evaluated on the same dataset (84.79) and outperforms the centralized setting (81.74) when uncertainty is utilized to enhance predictions. 
This improvement is especially high in scenarios with limited data, such as Learn2Reg~\cite{zhoubing2016l2r}, where overfitting occurs in the non-federated case (86.35 vs. 74.36).
The models trained on one dataset generally performed poorly when applied to other datasets (49.71). 
In the centralized case, the predictions could not be enhanced by performing uncertainty weighted averaging (81.74 vs. 79.40).
This could be explained by a worse calibration of the central trained model in terms of ECE (12.42 vs. 16.54, see Fig.~\ref{fig:ece-and-improvement}).

\section{Discussion}

The information for segmenting all structures across the datasets is encoded in the (federated trained) backbone.
Unable to assign a previously unseen label in its fixed number of channels, the segmentation head only learns the presence of some structures encoded in the uncertainty, without being explicitly trained for it. 
For example, Fig.\ref{fig:qualitative-results} illustrates how a network with a head trained on the BCV Cervix\cite{landman2015bcv} dataset can segment organs well in the uncertainty when applied to the AMOS~\cite{yuanfeng2022amos} dataset, which contains none of the labels used for training.
In federated learning, the segmentation heads with non-overlapping labels may use the same feature maps from the backbone, whereas in the centralized case, the heads tend to use different feature maps.
The improvement of our proposed FUNAvg is larger for underrepresented labels (see Fig.~\ref{fig:ece-and-improvement} and Tab.~\ref{tab:results-per-organ} in the Appendix). 
We believe that predictions for these structures are encoded in the uncertainties of the other clients, whereas for labels present at many sites, predictions are already encoded in the output. 
For these labels, our proposed method may lead to slightly worse results. 
Future work could explore whether uncertainty can be used more targeted during averaging in such cases.

\begin{credits}
\subsubsection{\ackname} The work was done during a research stay of MT at Menze lab at University Zurich (UZH) with support from the DAAD (German Academic Exchange Service). 
MT and SE are supported by grants from the Klaus Tschira Foundation within the Informatics for Life framework, by the DZHK and BMBF, in particular BMBF-SWAG Project 01KD2215D.
FN and BM are supported by the Helmut Horten foundation.
The authors gratefully acknowledge the data storage service SDS@hd  supported by MWK and DFG through grant INST 35/1314-1 FUGG and INST 35/1503-1 FUGG.

\subsubsection{\discintname}
The authors have no competing interests to declare that are
relevant to the content of this article.
\end{credits}

\bibliographystyle{splncs04}
\bibliography{Paper-1396}

\newpage

\section*{Appendix}

\begin{table}[H]
    \caption{Comparison of DICE scores per segmented structure. We perform an evaluation for each structure and compare the same methods as in Tab.~\ref{tab:results}. \textbf{M} denotes the row-wise mean. The full names of the segmented structures can be found in Fig.~\ref{fig:ece-and-improvement}.}
    \begin{tabular}{
      p{0.12\textwidth} |
      C{0.059\textwidth}
      C{0.059\textwidth} 
      C{0.059\textwidth} 
      C{0.059\textwidth} 
      C{0.059\textwidth} 
      C{0.059\textwidth} 
      C{0.059\textwidth} 
      C{0.059\textwidth} 
      C{0.059\textwidth}
      C{0.059\textwidth}
      C{0.059\textwidth}
      C{0.059\textwidth} |
      C{0.059\textwidth}
    }
    \toprule
     & \textbf{A} & \textbf{DD} & \textbf{Eso} & \textbf{GB} & \textbf{LK} & \textbf{L} & \textbf{P} & \textbf{RK} & \textbf{Spl} & \textbf{Sto} & \textbf{Tra} & \textbf{UB} & \textbf{M} \\
    \midrule
    Same & 88.7 & \textbf{61.2} & 74.0 & \textbf{75.8} & \textbf{88.8} & \textbf{92.6} & 60.9 & \textbf{89.1} & \textbf{90.6} & \textbf{78.9} & 84.0 & \textbf{78.7} & \textbf{80.3} \\
    Other & 81.8 & 38.1 & 68.0 & 40.9 & 75.6 & 49.2 & 39.2 & 71.7 & 61.4 & 68.4 & 76.5 & 40.2 & 59.3 \\
    \midrule
    CenAvg & 87.4 & 0.0 & 55.2 & 0.0 & 83.4 & 89.7 & 65.1 & 84.9 & 89.3 & 72.9 & 90.3 & \textbf{74.5} & 66.1 \\
    CUNA. & 82.4 & 0.0 & 74.1 & 0.0 & 83.0 & 86.8 & 61.8 & 83.8 & 86.4 & 73.3 & 87.7 & 70.8 & 65.8 \\
    FedAvg & \textbf{88.9} & 12.6 & 80.2 & 60.9 & 81.9 & 90.5 & \textbf{70.4} & 83.9 & 90.1 & 55.6 & 91.6 & 73.3 & 73.3 \\
    FUNA. & 88.1 & \textbf{51.3} & \textbf{81.7} & \textbf{73.5} & \textbf{85.3} & \textbf{91.3} & 66.2 & \textbf{86.5} & \textbf{90.2} & \textbf{78.7} & \textbf{92.8} & 71.8 & \textbf{79.8} \\
    \bottomrule
\end{tabular}
\label{tab:results-per-organ}
\end{table}

\begin{figure}[H]
    \centering
    \centering
    \includegraphics[width=0.81\textwidth]{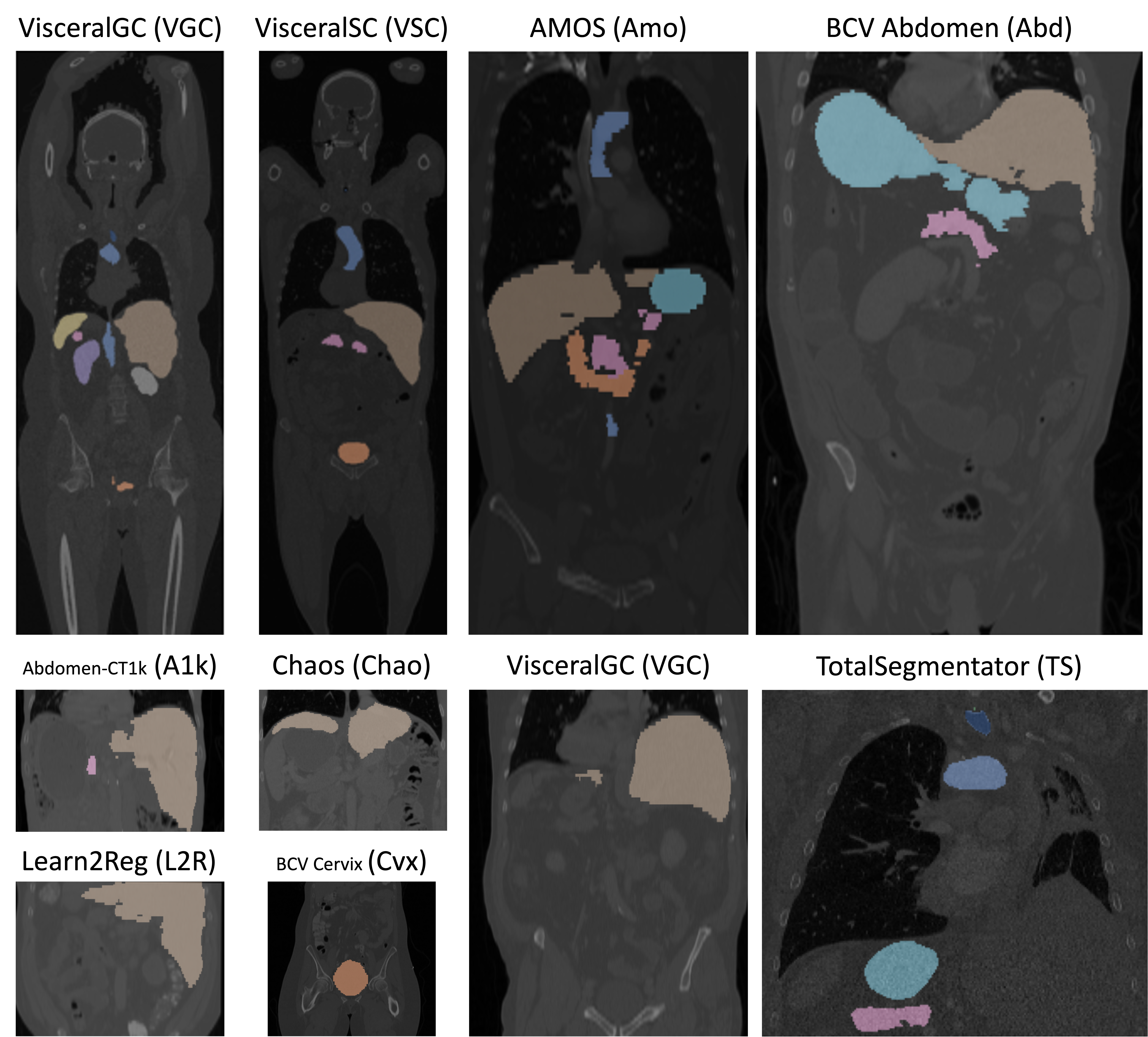}
    \caption{One representative slide from each dataset used in the federation with corresponding labels. Additional to the different annotated structures in each dataset, the field of view and with that the intensity distributions vary across datasets.}
    \label{fig:qualitative-comprison-datasets}
\end{figure}

\begin{table}[H]
    \caption{Comparison of DICE scores per segmented structure and dataset.  The full names of the segmented structures can be found in Fig.~\ref{fig:ece-and-improvement}.}
    \begin{tabular}{
      p{0.08\textwidth} 
      p{0.08\textwidth} |
      C{0.062\textwidth}
      C{0.062\textwidth} 
      C{0.062\textwidth} 
      C{0.062\textwidth} 
      C{0.062\textwidth} 
      C{0.062\textwidth} 
      C{0.062\textwidth} 
      C{0.062\textwidth} 
      C{0.062\textwidth}
      C{0.062\textwidth}
      C{0.062\textwidth}
      C{0.062\textwidth}
    }
    \toprule
     \textbf{DS} & \textbf{Meth.} & \textbf{A} & \textbf{DD} & \textbf{Eso} & \textbf{GB} & \textbf{LK} & \textbf{L} & \textbf{P} & \textbf{RK} & \textbf{Spl} & \textbf{Sto} & \textbf{Tra} & \textbf{UB} \\
    \midrule
    VSC & Cen & 80.44 & - & - & - & 79.36 & 89.23 & 42.77 & 84.44 & 85.91 & - & 92.57 & 65.45 \\
     & CUN & 73.53 & - & - & - & 80.73 & 84.91 & 39.11 & 84.59 & 81.75 & - & 88.32 & 63.18 \\
     & Fed & 83.05 & - & - & - & 78.47 & 90.86 & 43.19 & 79.36 & 87.92 & - & 93.82 & 64.7 \\
     & FUN & 80.02 & - & - & - & 80.38 & 90.49 & 39.94 & 83.28 & 86.86 & - & 92.15 & 64.99 \\
    \midrule
    LiTS & Cen & - & - & - & - & - & 91.24 & - & - & - & - & - & - \\
     & CUN & - & - & - & - & - & 87.95 & - & - & - & - & - & - \\
     & Fed & - & - & - & - & - & 92.51 & - & - & - & - & - & - \\
     & FUN & - & - & - & - & - & 93.06 & - & - & - & - & - & - \\
    \midrule
    Abd & Cen & 91.31 & - & - & - & 91.67 & 90.35 & 64.5 & 89.61 & 92.03 & 72.47 & - & - \\
     & CUN & 85.82 & - & - & - & 89.65 & 87.39 & 62.01 & 88.92 & 90.46 & 72.45 & - & - \\
     & Fed & 91.5 & - & - & - & 88.98 & 92.87 & 74.01 & 91.3 & 93.36 & 57.5 & - & - \\
     & FUN & 91.9 & - & - & - & 91.59 & 92.67 & 67.71 & 91.96 & 94.35 & 77.94 & - & - \\
    \midrule
    Cvx & Cen & - & - & - & - & - & - & - & - & - & - & - & 85.05 \\
     & CUN & - & - & - & - & - & - & - & - & - & - & - & 80.4 \\
     & Fed & - & - & - & - & - & - & - & - & - & - & - & 78.25 \\
     & FUN & - & - & - & - & - & - & - & - & - & - & - & 78.88 \\
    \midrule
    Chao & Cen & - & - & - & - & - & 95.41 & - & - & - & - & - & - \\
     & CUN & - & - & - & - & - & 92.86 & - & - & - & - & - & - \\
     & Fed & - & - & - & - & - & 91.92 & - & - & - & - & - & - \\
     & FUN & - & - & - & - & - & 94.63 & - & - & - & - & - & - \\
    \midrule
    Abd1k & Cen & - & - & - & - & - & 89.62 & 68.5 & - & 90.81 & - & - & - \\
     & CUN & - & - & - & - & - & 86.45 & 66.4 & - & 87.64 & - & - & - \\
     & Fed & - & - & - & - & - & 91.81 & 78.91 & - & 92.46 & - & - & - \\
     & FUN & - & - & - & - & - & 91.7 & 75.74 & - & 92.46 & - & - & - \\
    \midrule
    L2R & Cen & - & - & - & - & - & 88.15 & - & - & 87.75 & - & - & - \\
     & CUN & - & - & - & - & - & 86.88 & - & - & 84.94 & - & - & - \\
     & Fed & - & - & - & - & - & 86.49 & - & - & 83.73 & - & - & - \\
     & FUN & - & - & - & - & - & 89.26 & - & - & 83.44 & - & - & - \\
    \midrule
    TS & Cen & 93.36 & 0.0 & 64.99 & 0.0 & 86.24 & 87.65 & 76.42 & 93.07 & 93.6 & 80.42 & 86.01 & 74.01 \\
     & CUN & 89.41 & 0.0 & 81.3 & 0.0 & 85.53 & 84.74 & 71.74 & 89.77 & 89.92 & 78.44 & 85.51 & 72.47 \\
     & Fed & 93.59 & 17.48 & 88.27 & 66.61 & 83.2 & 89.78 & 81.29 & 91.04 & 94.18 & 67.13 & 87.31 & 75.44 \\
     & FUN & 93.06 & 65.7 & 87.04 & 80.58 & 89.59 & 90.7 & 75.91 & 90.19 & 94.75 & 88.22 & 93.24 & 73.27 \\
    \midrule
    VGC & Cen & 86.44 & - & - & - & 91.65 & 93.53 & 76.02 & 92.82 & 93.44 & - & 92.25 & 73.39 \\
     & CUN & 78.32 & - & - & - & 90.36 & 90.37 & 69.77 & 90.05 & 90.08 & - & 89.34 & 67.19 \\
     & Fed & 88.37 & - & - & - & 81.33 & 92.55 & 74.09 & 81.77 & 94.35 & - & 93.54 & 74.94 \\
     & FUN & 86.03 & - & - & - & 86.52 & 93.67 & 70.88 & 88.64 & 93.54 & - & 93.04 & 70.13 \\
    \midrule
    AMOS & Cen & 85.25 & 0.0 & 45.48 & 0.0 & 68.32 & 82.19 & 62.4 & 64.47 & 81.87 & 65.74 & - & - \\
     & CUN & 84.68 & 0.0 & 66.93 & 0.0 & 68.71 & 79.56 & 61.63 & 65.75 & 79.88 & 68.87 & - & - \\
     & Fed & 88.09 & 7.64 & 72.22 & 55.09 & 77.29 & 85.31 & 70.69 & 76.19 & 84.67 & 42.23 & - & - \\
     & FUN & 89.26 & 36.8 & 76.27 & 66.34 & 78.61 & 85.93 & 67.17 & 78.35 & 86.34 & 69.83 & - & - \\
    \bottomrule
\end{tabular}
\label{tab:results-per-organ-and-dataset}
\end{table}

\end{document}